\documentclass{article}

    \usepackage[preprint,nonatbib]{neurips_2020}

\usepackage[utf8]{inputenc} 
\usepackage[T1]{fontenc}    
\usepackage{hyperref}       
\usepackage{url}            
\usepackage{booktabs}       
\usepackage{amsfonts}       
\usepackage{nicefrac}       
\usepackage{microtype}      
\usepackage{xcolor}
\hypersetup{
    colorlinks=true,
    linkcolor=blue,
    filecolor=magenta,      
    urlcolor=black,
    citecolor=blue,
}

\usepackage{microtype}
\usepackage{graphicx}
\usepackage{subfigure}
\usepackage{booktabs} 

\usepackage{amsmath,amsfonts,bm}

\def\eqref#1{equation~\ref{#1}}

\def\1{\bm{1}}

\DeclareMathAlphabet{\mathsfit}{\encodingdefault}{\sfdefault}{m}{sl}
\SetMathAlphabet{\mathsfit}{bold}{\encodingdefault}{\sfdefault}{bx}{n}

\DeclareMathOperator*{\argmax}{arg\,max}

\usepackage{fancyref}
\usepackage{amsmath}
\graphicspath{{assets/}} 
\DeclareMathAlphabet\mathcalbf{OMS}{cmsy}{b}{n}
\usepackage{mathabx}
\usepackage{floatrow}
\usepackage{multirow}
\usepackage{cite}

\title{Enhanced Scene Specificity with Sparse Dynamic Value Estimation}

\author{%
  Jaskirat Singh \& Liang Zheng\\
  College of Engineering and Computer Science\\
  Australian National University\\
  Canberra, Australia \\
  \texttt{jaskirat.singh,liang.zheng@anu.edu.au} \\
}

\begin{document}

\maketitle

\begin{abstract}
Multi-scene reinforcement learning involves training the RL agent across multiple scenes / levels from the same task, and has become essential for many generalization applications. However, the inclusion of multiple scenes leads to an increase in sample variance for policy gradient computations, often resulting in suboptimal performance with the direct application of traditional methods (e.g. PPO, A3C). One strategy for variance reduction is to consider each scene as a distinct Markov decision process (MDP) and learn a joint value function dependent on both state $s$ and MDP $\mathcal{M}$. However, this is non-trivial as the agent is usually unaware of the 
underlying level at train / test times in multi-scene RL. Recently, Singh \emph{et al.} \cite{singh2020dynamic} tried to address this by proposing a dynamic value estimation approach that models the true joint value function distribution as a Gaussian mixture model (GMM). In this paper, we argue that the error between the true scene-specific value function $V(s,\mathcal{M})$ and the predicted dynamic estimate $\hat{V}(s,\mathcal{M})$ can be further reduced by progressively enforcing sparse cluster assignments once the agent has explored most of the state space. The resulting agents not only show significant improvements in the final reward score across a range of OpenAI ProcGen environments, but also exhibit increased navigation efficiency while completing a game level.
\end{abstract}

\section{Introduction}
\label{introduction}
Training on environments comprising of multiple scenes / variations from the same domain task (\emph{e.g.} different levels from a video game), has become a powerful strategy for countering over-fitting in deep reinforcement learning \cite{cobbe2018quantifying,justesen2018illuminating,zhang2018dissection,zhang2018study,igl2019generalization,packer2018assessing,sadeghi2016cad2rl}. However, such an approach comes at the price of increased sample variance in policy gradient computations \cite{cobbe2019leveraging,song2019empirical}. The high variance necessitates using more samples \cite{schulman2015high}, and thus, training high performance agents on these environments \cite{cobbe2019leveraging,kolve2017ai2,nichol2018gotta,juliani2019obstacle,beattie1612deepmind} invariably involves increasing the sample size per update step through the use of multiple parallel actors \cite{mirowski2016learning,wortsman2019learning,mnih2016asynchronous}. While parallel sample collection helps in stabilizing the learning process, the obvious disadvantages of lower sample efficiency and higher hardware constraints, suggest the need for specialized variance reduction techniques in multi-scene RL.

One such strategy is to replace the traditionally used scene-generic value function $V(s)$ with a scene-specific estimate $V(s,\mathcal{M})$ while computing the advantage function \cite{singh2020dynamic}. However, in the absence of information about the operational scene at train / test times, learning the scene-specific value function presents a challenging problem. Recently, Singh \emph{et al.} \cite{singh2020dynamic} showed that while a fine estimation of the joint value function is not feasible, a coarse approximation can be obtained by dividing the value function distribution into multiple clusters and then using episode trajectories to predict the assignment of the current state to each cluster (refer Section \ref{dve_revisit} for details).

\begin{figure}[ht]
\begin{center}
\centerline{\includegraphics[width=\linewidth]{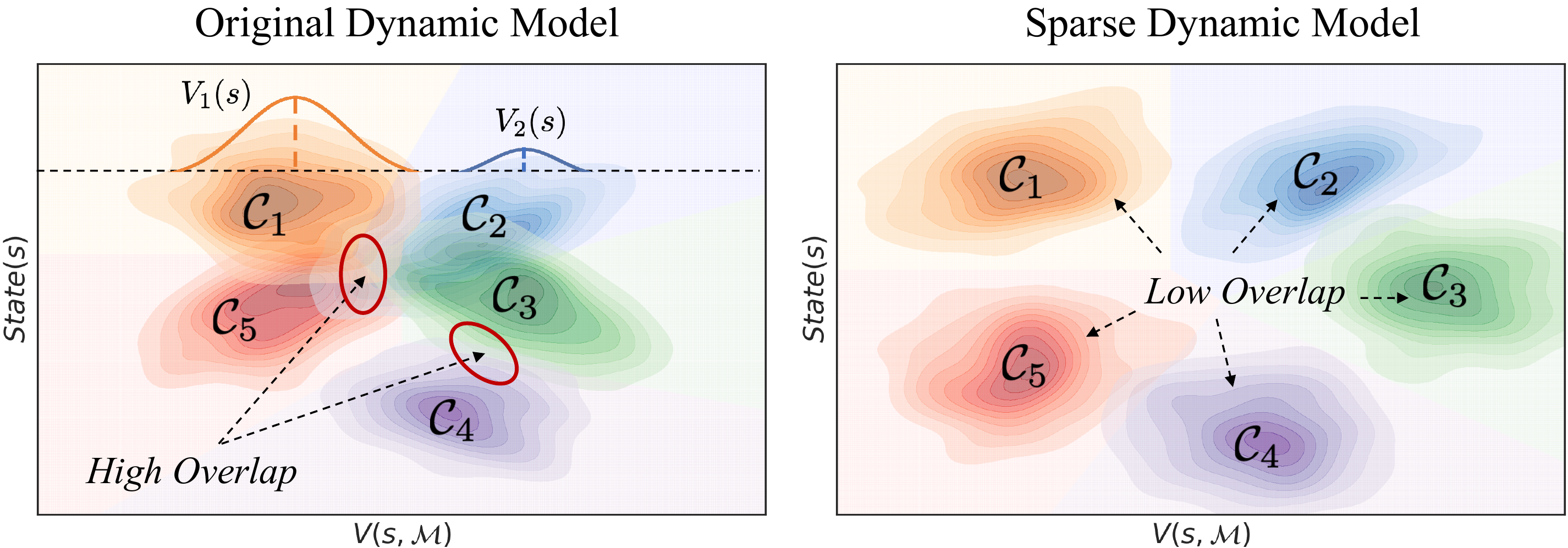}}
\caption{(\textbf{Left}) The clusters $\{\mathcal{C}_1,\mathcal{C}_2 ... \mathcal{C}_N\}$ originating from the original dynamic value estimation \cite{singh2020dynamic} can be approximated as multi-variate gaussian functions in $(s,V(s,\mathcal{M}))$ space. The overlap regions correspond to points of high confusion and are usually characterized by critical states / tricky obstacles \cite{singh2020dynamic}. We claim that such an overlap constitutes a waste of information available for joint value estimation and can be avoided by increasing the spread of learned cluster means (\textbf{Right}).}
\label{fig:overview}
\end{center}
\vskip -0.3in
\end{figure}

In this paper, we show that the scene-specificity of the dynamic value estimates can be further enhanced by enforcing sparse cluster assignments, once the agent has explored most of the state space and thus learned a good enough approximation of the cluster parameters. The sparse cluster probabilities along with application of traditional value function loss, have a combined effect of spreading out the learned clusters in $(s,V(s,\mathcal{M}))$ space.
We claim that the adjustment of dynamic clusters in this manner reduces the overall value function prediction error and support it with extensive testing on OpenAI ProcGen \cite{cobbe2019leveraging} environments. Fig. \ref{fig:overview} provides an overview of our method.

The main contributions of this paper are summarized as follows.
\begin{itemize}
    \item We introduce a novel \emph{confusion-contribution} loss for improving dynamic value estimation (DVE) \cite{singh2020dynamic}. The proposed loss decreases the overlap between learned dynamic clusters by progressively enforcing sparse cluster assignments.
    \item We demonstrate that the sparse clusters divide the overall state space into distinct sets of game skills. The collection of these skills represents a curriculum that the agent must master for effective game play.
    \item By comparing the game level trajectories for the non-sparse and sparse dynamic models, we show that the high navigation efficiency of our method and its tendency to limit unnecessary exploration, presents an effective alternative to explicit reward-shaping \cite{laud2004theory,laud2003influence,ng1999policy}, for penalizing longer episode-lengths / reward-horizons in multi-scene reinforcement learning.
\end{itemize}

\section{Relevant Background}

\subsection{Problem Setup}
The multi-scene learning problem is characterized by a set of MDPs $\mathcalbf{M}: \{\mathcal{M}_1,\mathcal{M}_2 ... \mathcal{M}_N\}$. Each MDP $\mathcal{M}$ is defined by state space $\mathcal{S}_\mathcal{M}$, transition probabilities  $\mathcal{P}_\mathcal{M}(s_{t+1}|s_t,a_t)$, reward function $r_\mathcal{M}(s_t,a_t,s_{t+1})$, discount factor $\gamma$ and the common action space $\mathcal{A}$. The agent with policy $\pi(a|s)$ then interacts with a randomly chosen MDP to generate a trajectory $\tau: \{s_0,a_0,s_1,a_1, ... s_T\}$ with total discounted reward $\mathcal{R}_{\tau} = \sum_{t=0}^{T-1} \gamma^t r(s_t,a_t,s_{t+1}) $. We aim to learn a policy $\pi^*$ such that the expected reward over the tuples $(\mathcal{M},\tau)$ is maximized, \emph{i.e.}, $\pi^* = \argmax_\pi \mathbf{E}_{\tau,\mathcal{M}}\left[\mathcal{R}_{\tau,\mathcal{M}}\right]$.

\subsection{Revisiting Dynamic Value Estimation}
\label{dve_revisit}
Singh \emph{et al.} \cite{singh2020dynamic} show that the true value function distribution across different scenes resembles a Gaussian Mixture Model and thus can be divided into clusters. The main idea of dynamic value estimation is to enforce muti-modal distribution learning by modelling the scene-specific value function as weighted sum over the mean value estimates for these clusters. Mathematically,

\begin{align}
    \hat{V}(s_t,\mathcal{M}) = \sum_{i=1}^{N_b} \alpha_i(s_t,\tau^{t-}) \  \hat{V}_i(s_t)  \qquad s.t. \quad \alpha_i > 0, \quad \sum_i^{N_b} \alpha_i=1, \label{eq:mean_sum}
\end{align}

where $\tau^{t-}$ is the trajectory till time $(t-1)$, $N_b$ is the number of clusters and $\alpha_i$, $\hat{V}_i(s)$ represent the cluster assignments and the value function mean for the $i^{th}$ cluster, respectively. 

From a qualitative perspective, \cite{singh2020dynamic} also show that the distribution of cluster assignments ($\alpha_i$) provides important intuition about the nature of states, and define two metrics to analyse the same, \emph{confusion} and \emph{contribution}.  Confusion ($\delta$) is a measure of uncertainty as to which cluster, the current state-trajectory pair $\{s_t,\tau^{t-}\}$ belongs to. On the other hand, contribution ($\rho$), as the name suggests, determines the `contribution' of a cluster in the overall value function estimation across  a general trajectory sequence $\tau: \{s_0,a_0,s_1,a_1, ... s_T\}$. Formally, confusion and contribution are defined as,
\begin{equation}
    \delta(s_t,\tau^{t-}) = \frac{1}{N_b . \sum_i \alpha^2_i(s_t,\tau^{t-})}, \qquad
    \rho_i(\tau) = \frac{1}{T} \sum_{t=1}^{T} \delta(s_t,\tau^{t-}) \  \alpha_i(s_t,\tau^{t-}).
    \label{eq:conf_contrib}
\end{equation}

\section{Motivation}
\label{motivation}
\subsection{Minimizing Cluster Overlap}
\label{minimize_cluster_overlap}
\begin{figure}[ht]
\vskip 0.1in
\begin{center}
\centerline{\includegraphics[width=\linewidth]{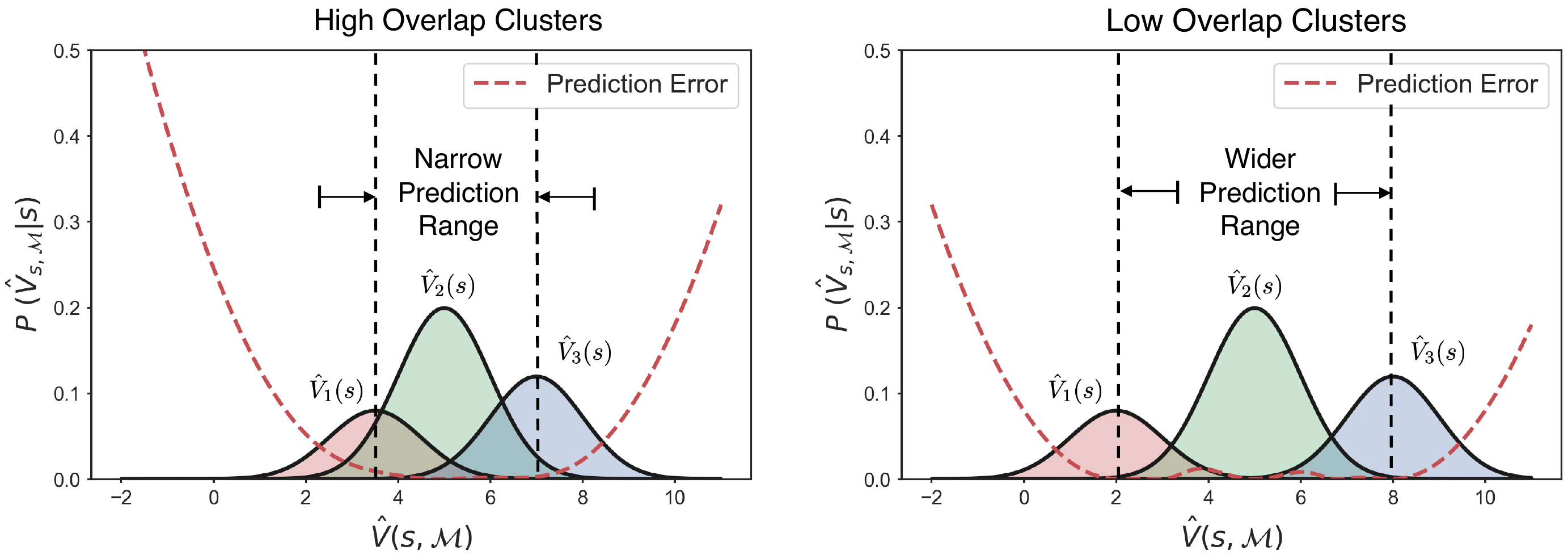}}
\caption{\textbf{Qualitative Analysis.} The dynamic value prediction model from Eq. \ref{eq:mean_sum} can be interpreted as the interpolation (using $\alpha_i$) across the learned cluster means $\hat{V}_i(s)$. Thus, as seen above, the prediction error is usually low in the range covered by the cluster centers. The original dynamic clusters with their high overlap are similar to the distribution shown on left, and have a very narrow range of low prediction error. As shown on the right, this region of low prediction error can be expanded by increasing the spread of learned dynamic cluster means.}
\label{fig:overlap}
\end{center}
\vskip -0.2in
\end{figure}

As shown in Fig. \ref{fig:overview}, we note that the original dynamic model leads to clusters with high overlap (high confusion) at critical states \cite{singh2020dynamic}. The high confusion states are usually characterized by presence of tricky obstacles / scenarios and are critical to the final episode reward. Given the value estimation model from Eq. \ref{eq:mean_sum}, it is understandable that the use mean squared error critic loss drives multiple cluster centers towards the true value of these critical states. However, such a behavior is undesirable as it reduces the range of value estimates $\hat{V}(s,\mathcal{M})$ covered through interpolation among cluster means in Eq. \ref{eq:mean_sum}. Fig. \ref{fig:overlap} explains how the spread of dynamic cluster means affects the prediction error $||V(s,\mathcal{M})-\hat{V}(s,\mathcal{M})||$ across $\mathcal{M} \in \mathcalbf{M}$. Consequently, we conjecture that the overall prediction error can be reduced by minimizing the overlap between the learned dynamic clusters.

\subsection{Correlation Analysis}
\label{corr_analysis}
\begin{figure}[ht]
\floatbox[{\capbeside\thisfloatsetup{capbesideposition={right,center},capbesidewidth=7cm}}]{figure}[\FBwidth]
{\caption{Results showing Pearson correlation \cite{benesty2009pearson} coefficient $(r)$ between inverse confusion $(1/\delta)$ and total reward score $(\mathcal{R})$. For most games the correlation coefficient is greater than 0.5, which points to the statistical significance of the analysis done in section \ref{minimize_cluster_overlap}. We next demonstrate how the original dynamic training can be modified to achieve lower confusion in cluster assignments.}\label{fig:correlation}}
{\includegraphics[width=7cm]{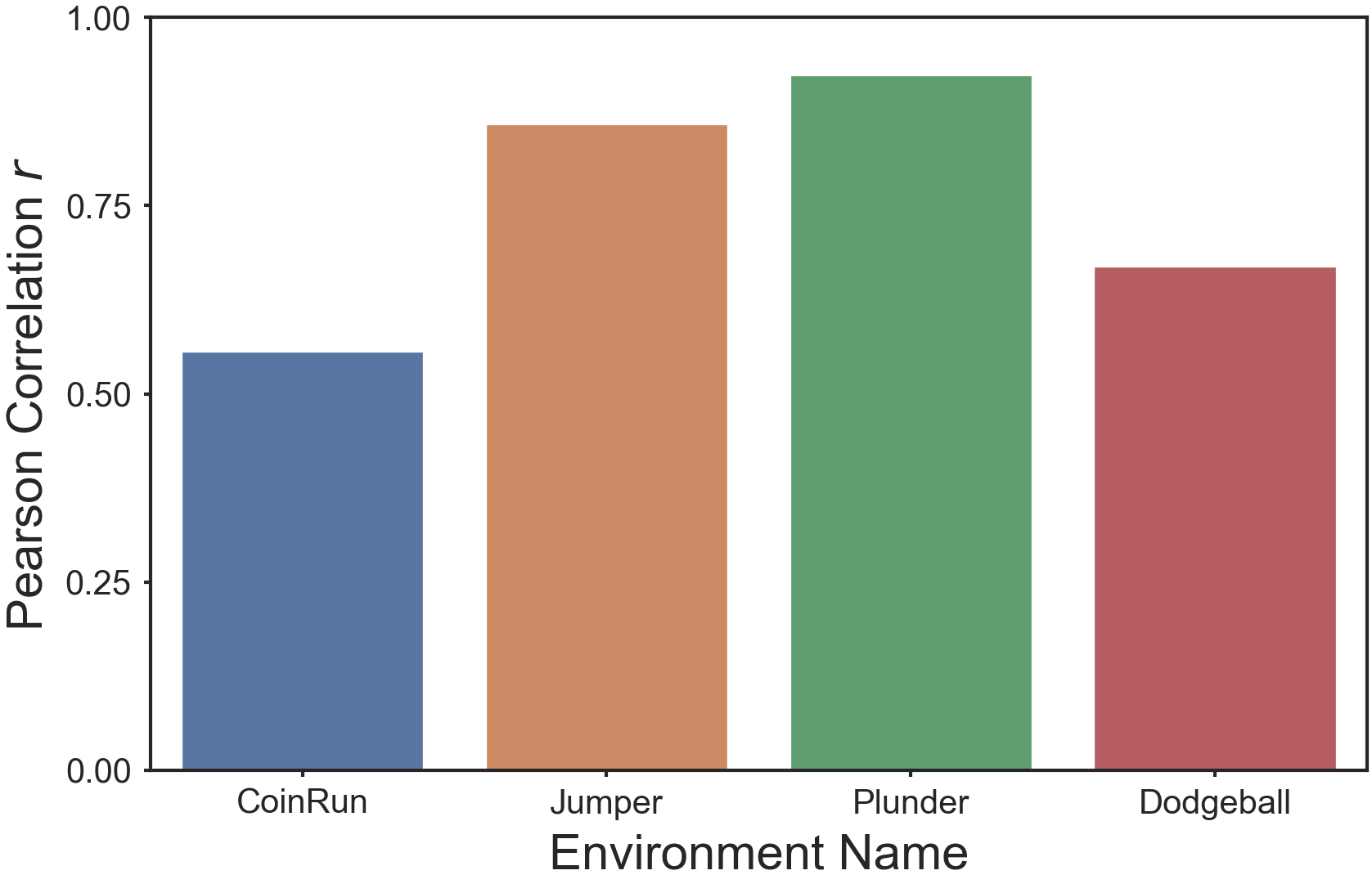}}
\end{figure}

As the clusters move far apart from each other, the cluster assignments $\alpha_i$ for a given tuple $(s,\tau^-)$ tend towards a one-hot encoding, with the one corresponding to the closest cluster. This implies that a higher spread in cluster means corresponds to a sparser cluster assignment distribution and can be measured using the confusion $\delta$ (refer Eq. \ref{eq:conf_contrib}).
Hence, to test the initial validity of the above analysis, we compute the correlation between final model performance and inverse confusion ($1/\delta$), while training on OpenAI's ProcGen \cite{cobbe2019leveraging} environments. The samples for this testing are collected randomly during the first 50M timesteps of training across 4 distinct runs with the original dynamic model. The Pearson correlation \cite{benesty2009pearson} coefficients  for various ProcGen games are shown in Fig. \ref{fig:correlation}. The results clearly corroborate our analysis from section \ref{minimize_cluster_overlap} and show a high correlation between reduced confusion and improved model performance.

\section{Sparse Dynamic Value Estimation}
\label{our_method}
\begin{figure}[ht]
\begin{center}
\centerline{\includegraphics[width=\linewidth]{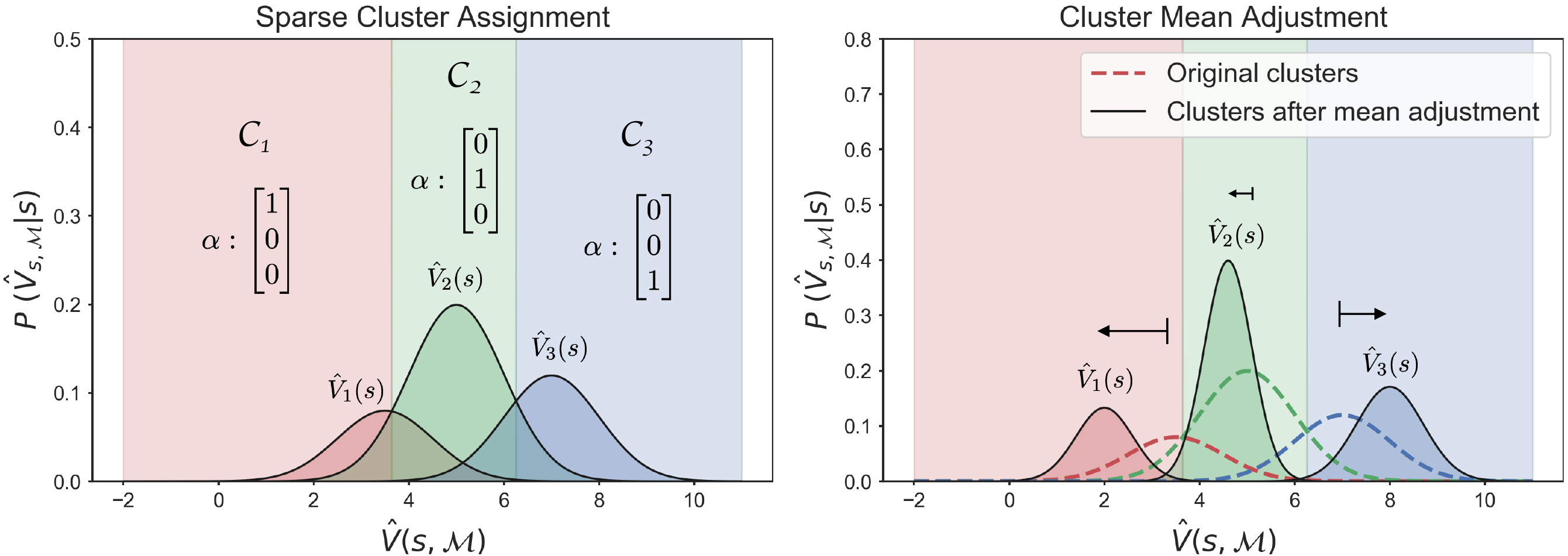}}
\caption{\textbf{Illustration.} Example showing how sparse cluster assignments help in reducing the overlap between clusters learned by the dynamic model. (\textbf{Left}) Each point $(s,\mathcal{M})$ is allocated to the most probable cluster based on the cluster assignments $\alpha_i$. (\textbf{Right}) The means of each cluster adjust to reflect the expected value estimate of all $(s,\mathcal{M})$ pairs in the modified cluster assignments. Note that the hard sparse assignment is for illustration purposes only. In practice, the sparsity is introduced progressively as the new cluster means are learned through the value function loss.}
\label{fig:sparse_steps}
\end{center}
\vskip -0.1in
\end{figure}

Given the analysis from Section \ref{corr_analysis}, we note that increasing the inter-cluster mean variance leads to sparser cluster assignment distribution. We claim that the reverse is also true, \emph{i.e.}, an appropriate spread in learned cluster means can be obtained by progressively enforcing sparse cluster assignments followed by adjustment of cluster means. Fig. \ref{fig:sparse_steps} illustrates this process on a sample GMM distribution.

\subsection{Enforcing Sparsity}

The sparsity condition is equivalent to maximization of the L2 norm for cluster assignments $\{\alpha_1,\alpha_2, \dots \alpha_{N_b}\}$ and thus using Eq. \ref{eq:conf_contrib}, corresponds to minimal confusion ($\delta$). However, we note that a mere enforcement of sparsity may encourage convergence to solutions where only one of the clusters is active. We also want to ensure that each cluster contributes equally in the $(s,\mathcal{M})$ space. To achieve this, we propose the following \emph{confusion-contribution loss},
\begin{equation}
    L^{CC} = k_1 \ \mathbf{E}_{s_t,\tau^{t-}} \left[\log \delta(s_t,\tau^{t-})\right] + k_2 \ \mathbf{E}_{\tau}  \left[\log \left(\sum_i^{N_b} \rho^2_i(\tau)\right)\right].
\end{equation}

We must emphasize that the state space must have already been well explored by the agent, prior to the application of confusion-contribution loss. If applied prematurely, due to the continuous nature of neural networks, the sparse cluster assignment is incorrectly generalized across the entire state space. This would lead to a detrimental impact on value function estimation for the currently unexplored states. Also, such a mistake is hard to recover from, because for any state $s$, the sparse assignment ensures that the gradients for all but one cluster are approximately zero.

\section{Evaluation on OpenAI ProcGen}
\subsection{Experimental Design}

\textbf{Training Details.} The network design for the dynamic model is quite similar to the one described in \cite{singh2020dynamic}. The states are fed through an IMPALA-CNN \cite{espeholt2018impala} + LSTM \cite{duan2016rl} network to output a joint latent representation, used for learning both the policy and the value function. The critic network uses these latent representations to predict cluster assignments $\alpha_i$ and mean value estimates $\hat{V}_i(s)$. Finally, the predicted value function $\hat{V}(s,\mathcal{M})$ is computed using Eq. \ref{eq:mean_sum}. Similar to \cite{cobbe2019leveraging}, the agent is trained using Proximal policy optimization (PPO) \cite{schulman2017proximal} with 4 parallel workers. The only point of difference with the original dynamic model is the application of confusion-contribution loss $(L_{CC})$ at a suitable stage in the training process. The loss coefficients $(k_1,k_2)$ determine the balance between confusion and contribution, and are chosen through extensive hyper-parameter search for each environment.

We test our method on 8 ProcGen \cite{cobbe2019leveraging} environments: CoinRun, CaveFlyer, Climber, Jumper, Plunder, Dodgeball, FruitBot and StarPilot. Note that each game is characterized by a different rate of state exploration and training trajectories. Thus, depending upon the type of environment, we adopt the following strategies for obtaining sparse boosts.

\textbf{Pre-boost.} For games allowing rapid state space exploration at the beginning, the confusion-contribution loss can be applied quite early to promote sparsity. In fact, because the policy gradient and value function loss dominate the initial training updates, we apply the confusion-contribution loss from the start. However, the coefficients $(k_1,k_2)$ are kept moderately small so as to encourage the network to progressively converge to a sparse cluster assignment over the first quarter timesteps. CoinRun, CaveFlyer, Climber and Jumper belong to this set and are labelled as \emph{class-1} environments.

\textbf{Post-boost.} In contrast, other games display a much more gradual expansion of explored state space, exhibiting a positive correlation between episode lengths and the total reward. Sparse-boosting for such environments, can only be applied after the rate of increase of average episode length has declined. Thus, the application of confusion-contribution loss is usually preceded by pre-training with the original dynamic model for 50M timesteps (per worker). Games like Plunder, Dodgeball, FruitBot and StarPilot are part of this set and are labelled as \emph{class-2} environments.

We also train the vanilla-LSTM based RL$^2$ \cite{duan2016rl} and non-sparse dynamic models from \cite{singh2020dynamic}, to show a comprehensive comparison between model performances. For consistency reasons, a pre-training procedure same as the one described above is followed, for all class-2 environments. All results are reported as an average across 4 distinct runs using 500 levels for training.

\subsection{Results}
\begin{figure}[ht]
\begin{center}
\centerline{\includegraphics[width=0.97\linewidth]{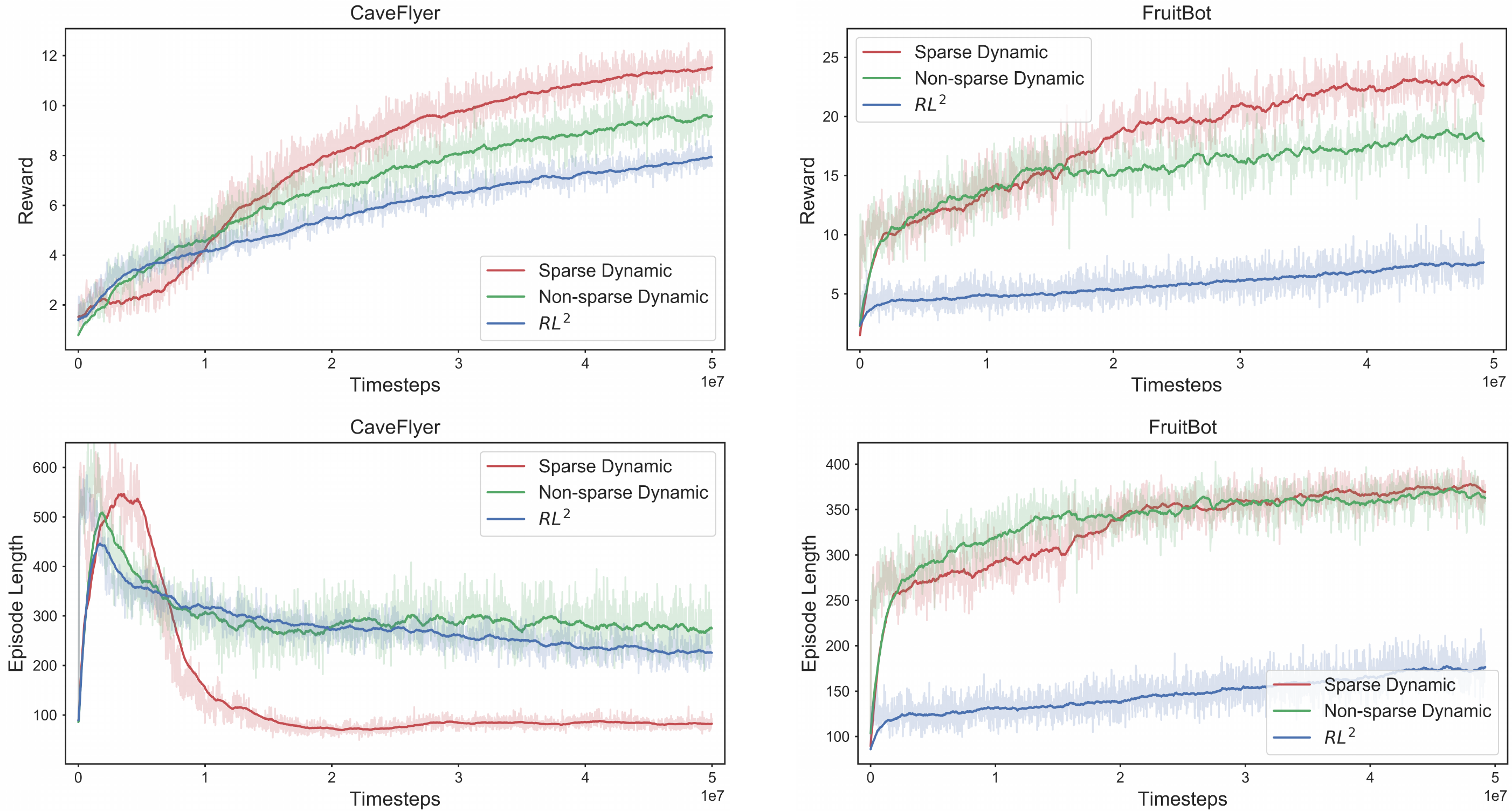}}
\caption{Learning curves for RL$^2$, sparse and non-sparse dynamic models, illustrating differences in sample efficiency, total reward and and episode lengths.}
\label{fig:procgen_train_curves}
\end{center}
\vskip -0.2in
\end{figure}

\textbf{Class-1.} The sparse model leads to consistent performance improvements in all 4 class-1 environments. Fig. \ref{fig:procgen_train_curves} shows the total reward and average episode length curves during the training process for the Caveflyer environment. We clearly see that sparse training leads to significant gains in both final reward and sample efficiency over the regular dynamic model. For instance, we report an increase of 28.3\% and 22.4\% in the final episode reward, over the non-sparse dynamic model, for the games of CaveFlyer and Climber respectively (refer Table \ref{tb:procgen_results}).

Furthermore, as shown in Fig. \ref{fig:procgen_train_curves} and Table \ref{tb:procgen_results}, the sparse model leads to better reward scores while on average, using much fewer timesteps per episode\footnote{Note that the ProcGen environments have no explicit penalty for longer episode lengths.}. We call this phenomenon as enhanced navigation efficiency and delve into it in detail in Section \ref{navigation_efficiency}.

\textbf{Class-2.} Fig. \ref{fig:procgen_train_curves} reports the results for the FruitBot (class-2) environment using the post-boost strategy. While the baseline RL$^2$ and non-sparse dynamic models show a saturation in model performance with the extended training protocol, the sparse model loss leads to continued gains in reward scores. Interestingly, we also see that a saturation in rate of state space exploration is necessary for getting gains with the sparse model. This is illustrated through the training curves for the game of FruitBot (refer Fig. \ref{fig:procgen_train_curves}), where relative gains over the non-sparse model occur only after a decline in the rate of increase of average episode length. 

For sake of completeness, we report the results for all class-1 and class-2 environments in Table \ref{tb:procgen_results}.

\begin{table}
  \vskip 0.1in
  \centering
  \begin{tabular}{c|l|ccc|ccc}
    \toprule
    \multicolumn{2}{c}{} & \multicolumn{3}{c}{Total Reward}  & \multicolumn{3}{c}{Episode Length}\\
    \midrule
    Class & Environment & RL$^2$ & DVE & Sparse DVE & RL$^2$ & DVE & Sparse DVE \\
    \midrule
    \multirow{4}{*}{Class 1} & CoinRun  & 7.75 & 9.16 & \textbf{9.62} & 126 & 78.99 & 62.1 \\
    & CaveFlyer  & 6.82 & 9.02 & \textbf{11.57}  & 225.6 & 275.1 & 75.2\\
    & Climber & 7.50 & 8.14 & \textbf{10.17} & 178.1	& 226.6	& 170.5 \\
    & Jumper  & 6.61 & 6.52 & \textbf{6.65} & 236.4 & 180.3 & 78.9\\
    \midrule
    \multirow{5}{*}{Class 2} & Plunder  & 7.13 & 17.16 & \textbf{18.42} & 495.1	& 780.6 & 739.4\\
    & DodgeBall & 10.98 & 11.25 & \textbf{12.76} & 401.5 & 459.9 & 285.8 \\
    & FruitBot  & 7.33 & 18.32 & \textbf{23.08} & 172.1 & 364.5 & 374.2\\
    & StarPilot & 17.94 & 18.08 & \textbf{19.81} & 327.6 & 342.9 & 320.2 \\
    \bottomrule
  \end{tabular}
  \caption{Performance comparison in final reward and average episode length for both class 1 and 2 environments. Our method achieves higher total rewards while needing much shorter episode lengths.}
  \label{tb:procgen_results}
\end{table}

\section{What are Clusters Made of?}
\label{cluster_state_analysis}

\begin{figure}[ht]
\begin{center}
\centerline{\includegraphics[width=\linewidth]{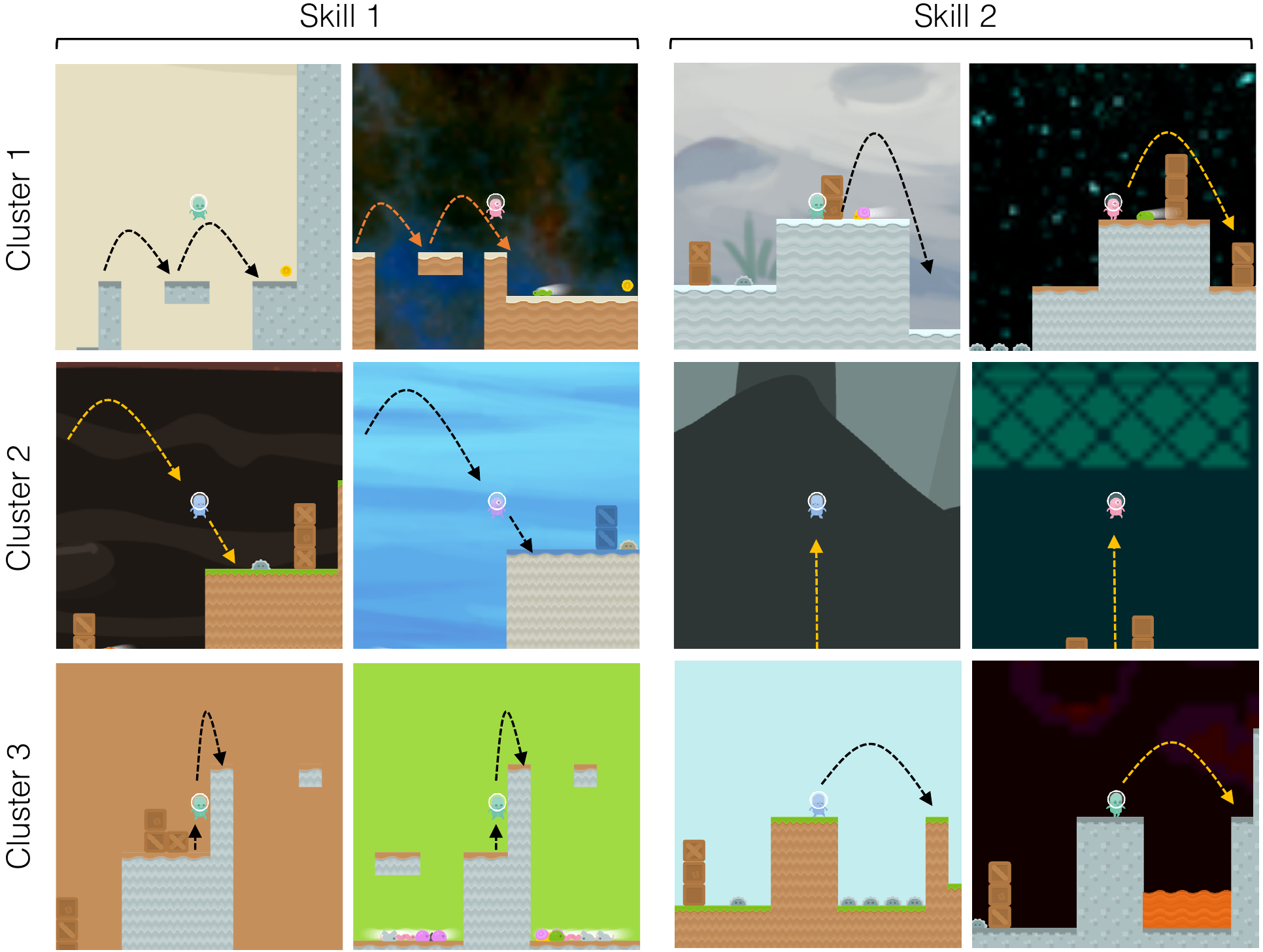}}
\caption{Examples of key obstacles types learned by the each cluster in the CoinRun Environment. We note that the sparse training divides the overall state space into a distinct sets of game skills.}
\label{fig:coinrun_cluster_feats}
\end{center}
\vskip -0.1in
\end{figure}

Given the non-sparse nature of original cluster probability distribution, Singh \emph{et al.} \cite{singh2020dynamic} use the normalized contribution scores as a measure of similarity between the basis MDPs \cite{singh2020dynamic} and a particular game level. However, such a comparison is not helpful in understanding the key features that differentiate each basis cluster. In this section, we use the sparse property of our method to visualize different obstacle types characteristic of each cluster in the CoinRun Environment.

To visualize the distinguishing features for each cluster, we first extract the set of states $\mathcal{S}_i$ for which each cluster is active. The latent representations (output of the LSTM network) for these states are used to map each $s \in \mathcal{S}_i$ to a two dimensional embedding space using TSNE \cite{maaten2008visualizing}. This embedding is then manually analysed for clusters to the identify the salient obstacle classes. 

Fig. \ref{fig:coinrun_cluster_feats} shows some key obstacle types for each cluster. We observe that each cluster is responsible for predicting the value function on a distinct set of obstacles / skills. For instance, cluster-1 is responsible for value estimation in cases like double-jump from one side to another (skill-1) and crossing over moving enemies (skill-2). On the other hand, cluster-2 handles landing after jumps from higher ground (skill-1) and high jumps with very limited visibility of the coming obstacles (skill-2). Finally, cluster-3 takes care of precision climbs (skill-1) and jumps over wide valleys (skill-2).

Thus, we see that each disjoint state space set $\mathcal{S}_i, \ i\in[1,N_b]$ represents a distinct curriculum of game skills that must be learned for mastering the overall multi-scene game environment. This division is analogous to human learning where it is quite common to break down a complex task into a set of manageable skills before attempting the complete task. 

\section{Navigation Efficiency}
\label{navigation_efficiency}
\begin{figure}[ht]
\begin{center}
\centerline{\includegraphics[width=\linewidth]{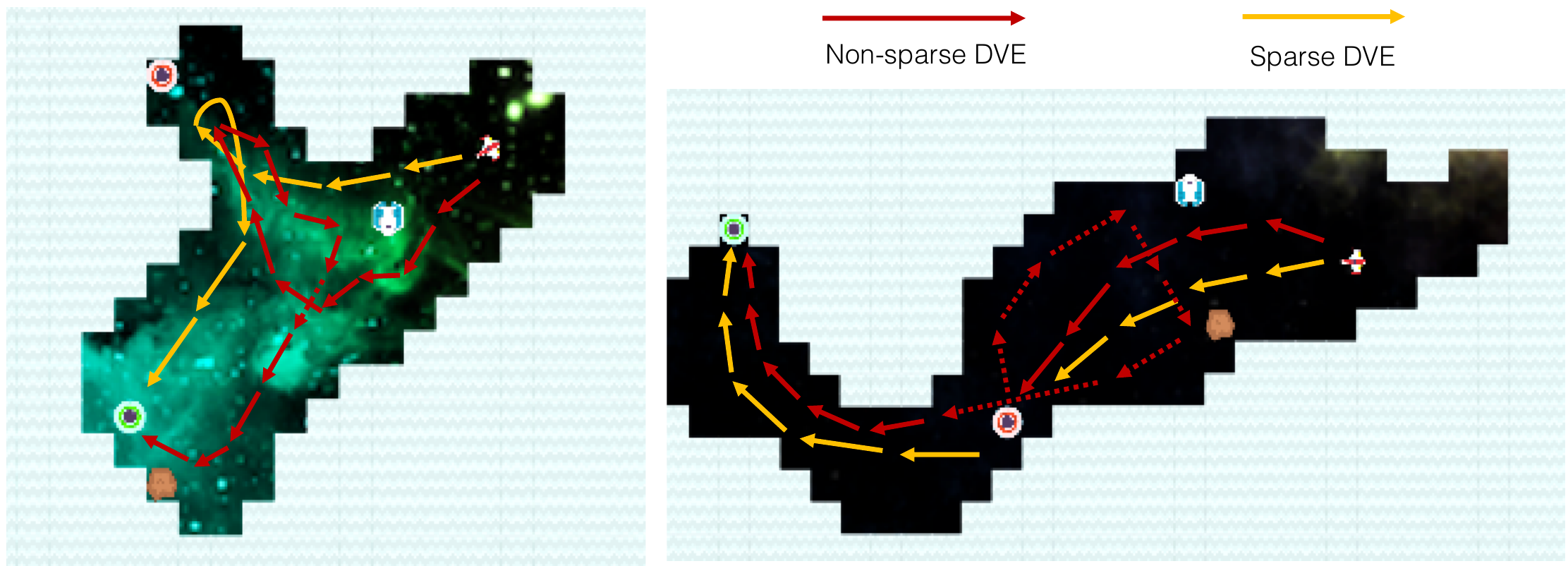}}
\caption{Demonstrating qualitative difference between trajectories for sparse and non-sparse dynamic agents. Our method shows higher efficiency in navigating to the final goals (red \& green spheres).}
\label{fig:trajectories}
\end{center}
\vskip -0.1in
\end{figure}

A peculiar feature arising as a result of applying sparse boosting can be seen in terms of improved \emph{navigation efficiency}. That is, the agent on average uses fewer time-steps per episode while achieving similar or better reward score. This massive difference in time-steps results from two reasons:
\begin{itemize}
    \item The tendency to use fewer time-steps is a direct consequence of optimizing the discounted reward function with $\gamma < 1$ \cite{schulman2015high}. As a result, the agent is incentivized to minimize the number of steps between the current state and the next reward. Hence, more accurate policy updates (lower sample variance) should lead to fewer timesteps.
    \item As explained in Section \ref{our_method}, the expansion in state space after application of confusion-contribution loss can lead to potential errors in value function estimation. Thus, the sparse dynamic agent learns to maximize the utilization of already explored state space. 
\end{itemize}

In this section, we will analyse the first point in greater detail. We first note that not all critical game states correspond to a high overlap region in the non-sparse dynamic model. At a critical state with low overlap (and possibly high value prediction error), the computation of suboptimal value function, can lead the agent to underestimate the \emph{advantage} of choosing an action leading to a faster route to the final destination / goal. We next try to identify these critical states by comparing episode trajectories for the sparse and non-sparse dynamic agents on the CaveFlyer environment.

\textbf{Game Description.} The goal of the Caveflyer environment is to destroy the red spheres and finally reach the green sphere while avoiding intermediate obstacles. The agent receives a small reward of +3 on destroying a red sphere and an end of episode reward of +10 on successfully reaching the green one. Direct collisions with an obstacle or the red sphere cause immediate episode termination.

Trajectories for both sparse and non-sparse dynamic agents are shown in Fig. \ref{fig:trajectories}. We see that the non-sparse agent after destruction of the red sphere (critical state), effectively restarts its search for the next target, while often revisiting already encountered states. In contrast, the sparse agent with its more accurate value estimates, realizes that the expected value for exploring unseen parts of the cave is much higher than revisiting previous states. Doing so not only helps the sparse agent in reaching the end goals much faster, but also eliminates the need for evading obstacles it has already crossed.

We also note that, the balance between the sparse model's reluctance towards state space expansion and maximization of total reward can be modulated through the coefficients of the confusion-contribution loss. In this regard, the high navigation efficiency of our method provides an effective alternative to designing explicit reward shaping \cite{laud2004theory} penalties for promoting reduced episode lengths.

\section{Conclusion}
This paper introduces a novel \emph{confusion-contribution} loss which improves the efficiency of the recently proposed dynamic value estimation method, by progressively learning sparser cluster assignments. The resulting dynamic clusters \emph{contribute} equally to the overall value function estimation and display minimal inter-cluster overlap. The proposed approach consistently outperforms the vanilla-LSTM based RL$^2$ and non-sparse dynamic models on a range of OpenAI ProcGen environments, while on average using much fewer timesteps per episode to complete a game level. Additionally, the sparse training divides the overall state space into disjoint subsets. We show that each subset focuses on a distinct set of game-skills, which draws a strong parallel with the human learning paradigm.

\section*{Broader Impact}
While this work is largely theoretical, we believe that in the long term, it will have major impact in the upcoming area of AI-inspired learning \cite{sadler2019game}. Recent years have seen the field of deep reinforcement learning demonstrate tremendous success in achieving super-human performance in complex game play. Deepmind's Alphazero \cite{silver2017mastering}, Alphastar\cite{vinyals2019grandmaster} and OpenAI's Dota-2 \cite{berner2019dota} are some salient examples. Each such milestone is followed by an increased public interest to analyse and break down the policy of the trained RL agent into a set of simple skills than can be consumed by a human learner \cite{sadler2019game,alphago_teach}. This process is often manual and involves painstaking analysis across hundreds of game runs. As shown in Section \ref{cluster_state_analysis}, our method does this automatically by dividing the possible game scenarios (states) into distinct sets of game skills. While each set can be composed of other mini-skills, the broad division achieved by our method promises great potential in the development of semi-automatic, AI-inspired teaching tools for human players.

\bibliography{sparse_est}
\bibliographystyle{ieeetr}

\end{document}